\documentclass[
]{ceurart}

\usepackage{listings}
\usepackage{tikz}
\usetikzlibrary{positioning, arrows.meta, decorations.pathreplacing, calc, fit}

\begin{document}

\copyrightyear{2026}
\copyrightclause{Copyright for this paper by its authors.
  Use permitted under Creative Commons License Attribution 4.0
  International (CC BY 4.0).}

\conference{Ital-IA 2026: 6th National Conference on Artificial Intelligence, organized by CINI, June 18-19, 2026, Rome, Italy}

\title{Parameter-Efficient Multi-View Proficiency Estimation: From Discriminative Classification to Generative Feedback}

\author[1]{Edoardo Bianchi}[%
orcid=0000-0002-0963-9543,
email=edbianchi@unibz.it,
]
\cormark[1]

\address[1]{Faculty of Engineering, Free University of Bozen-Bolzano, Bozen-Bolzano, Italy}

\author[1]{Antonio Liotta}[%
orcid=0000-0002-2773-4421,
email=antonio.liotta@unibz.it,
]

\cortext[1]{Corresponding author.}


\begin{abstract}
Estimating how well a person performs an action, rather than
which action is performed, is central to coaching, rehabilitation, and
talent identification. This task is challenging because proficiency is
encoded in subtle differences in timing, balance, body mechanics, and
execution, often distributed across multiple views and short temporal
events. We discuss three recent contributions to multi-view
proficiency estimation on Ego-Exo4D. SkillFormer introduces a
parameter-efficient discriminative architecture for selective
multi-view fusion; PATS improves temporal sampling by preserving
locally dense excerpts of fundamental movements; and ProfVLM
reformulates proficiency estimation as conditional language
generation, producing both a proficiency label and expert-style
feedback through a gated cross-view projector and a compact
language backbone. Together, these methods achieve
state-of-the-art accuracy on Ego-Exo4D with up to $20\times$ fewer
trainable parameters and up to $3\times$ fewer training epochs
than video-transformer baselines, while moving
from closed-set classification toward interpretable feedback
generation. These results highlight a shift toward efficient, multi-view systems
that combine selective fusion, proficiency-aware sampling, and
actionable generative feedback.
\end{abstract}

\begin{keywords}
  Proficiency Estimation \sep
  Action Quality Assessment \sep
  Vision-Language Models \sep
  Multi-View Video Understanding \sep
  Sports Analytics
\end{keywords}

\maketitle

\section{Introduction}
Action quality assessment (AQA) and proficiency estimation move beyond
action recognition by focusing on \emph{how well} an action is
performed. This requires modelling subtle differences between
executions of the same task, such as body mechanics, timing, balance,
and the consistency of fundamental movements~\cite{AQA_survey}. These
cues unfold over several seconds, often appear as micro-events that
uniform sampling fails to preserve, and are best captured from
multiple camera angles. Recent multi-view, expert-annotated datasets
such as Ego-Exo4D~\cite{egoexo4d} and BASKET~\cite{pan2025basket}
now enable data-driven approaches to this problem. However,
applications such as coaching, rehabilitation, motor learning, and
talent identification require interpretable, multi-view-aware systems
rather than classifiers returning a single label.

In this work, we discuss three of our recent contributions on the
Ego-Exo4D benchmark: SkillFormer~\cite{10.1117/12.3093974}, a
parameter-efficient multi-view discriminative architecture;
PATS~\cite{pats}, an architecture-agnostic temporal sampling strategy;
and ProfVLM~\cite{BIANCHI2026104749}, the first vision--language
model to jointly generate a proficiency label and expert-style
commentary. An earlier work, Gate-Shift-Fuse~\cite{gsfmeccano},
provides context on the role of multimodal fusion. We describe the
architectures, report empirical findings on Ego-Exo4D, and summarize
the design principles most relevant for future work.

\section{Background and Related Works}
Action quality assessment has evolved from hand-crafted scoring
pipelines to deep models built on pretrained video
encoders~\cite{AQA_survey}. The multitask formulation of Parmar and
Morris~\cite{parmar2019mtl} showed that auxiliary captions and class
labels can regularise the regression target, while natural-language
explanation has emerged only recently through prompt-guided multimodal
interaction~\cite{zhang2024nae}.

Expert-annotated multi-view datasets have shifted attention toward
the alignment and fusion of synchronised streams carrying
complementary cues about body kinematics, object interactions, and
the surrounding environment. Ego-Exo4D~\cite{egoexo4d} is central to
this setting: it pairs an egocentric stream with up to four
exocentric views across six skill domains and provides both
proficiency labels and free-form expert commentary. Related benchmarks
such as BASKET~\cite{pan2025basket} further highlight the growing
interest in fine-grained skill assessment, although they do not
include natural-language feedback. Complementary modalities, such as
heart rate from eye-tracking cameras~\cite{egoppg}, are also emerging
as auxiliary signals for proficiency estimation.

Multi-view proficiency estimation also builds on broader modelling
trends. Video transformers such as TimeSformer~\cite{timesformer}
capture long-range spatio-temporal dependencies, while
instruction-tuned VLMs~\cite{llava} and compact language models such
as SmolLM2~\cite{smollm2} enable structured textual feedback.
LoRA~\cite{lora} provides parameter-efficient adaptation, and agentic
video systems are beginning to appear~\cite{videoagent,tacticexpert};
however, coaching agents that adapt feedback across sessions remain
largely unaddressed.


\section{Methods}
\label{sec:methods}

\subsection{Benchmark: the EgoExo4D Dataset}
\label{sec:dataset}
The main contributions we report are evaluated on
Ego-Exo4D~\cite{egoexo4d}. We use the demonstrator proficiency subset,
which contains time-synchronised multi-view videos of people
performing skilled activities: one egocentric stream and up to four
static exocentric views per take. The subset covers six domains
(cooking, basketball, soccer, dancing, music, and bouldering) and
provides, for each take, a four-level proficiency label---Novice,
Early Expert, Intermediate Expert, or Late Expert---together with
free-form expert commentary. We follow the protocol introduced in
egoPPG~\cite{egoppg} and adopted by
SkillFormer~\cite{10.1117/12.3093974} and PATS~\cite{pats}: $10\%$
of the official training set is held out for validation, while the
official validation set is used for testing. We report top-1 accuracy
and, for ProfVLM~\cite{BIANCHI2026104749}, BERTScore, METEOR, and
ROUGE-L against the ground-truth commentary.

\subsection{Preliminary Work: From Multimodal Fusion to Multi-View Proficiency}
\label{sec:antecedents}
Our earlier work on egocentric action recognition in industrial
settings~\cite{gsfmeccano} provides a foundation for the multi-view
models discussed here. It showed that complementary modalities (RGB
and depth in that case) can improve over single-stream models when
explicitly fused rather than merely concatenated. The approach ranked
second in the MECCANO 2023 challenge, achieving $52.57\%$ top-1
accuracy. This result motivated the fusion-oriented view adopted in
SkillFormer and ProfVLM, where synchronised camera streams are treated
as complementary evidence to be aligned, weighted, and integrated.

\subsection{SkillFormer: Discriminative Multi-View Proficiency Estimation}
\label{sec:skillformer}
\begin{figure}[t]
  \centering
  \begin{minipage}[c]{0.49\linewidth}
    \centering
    \includegraphics[width=\linewidth]{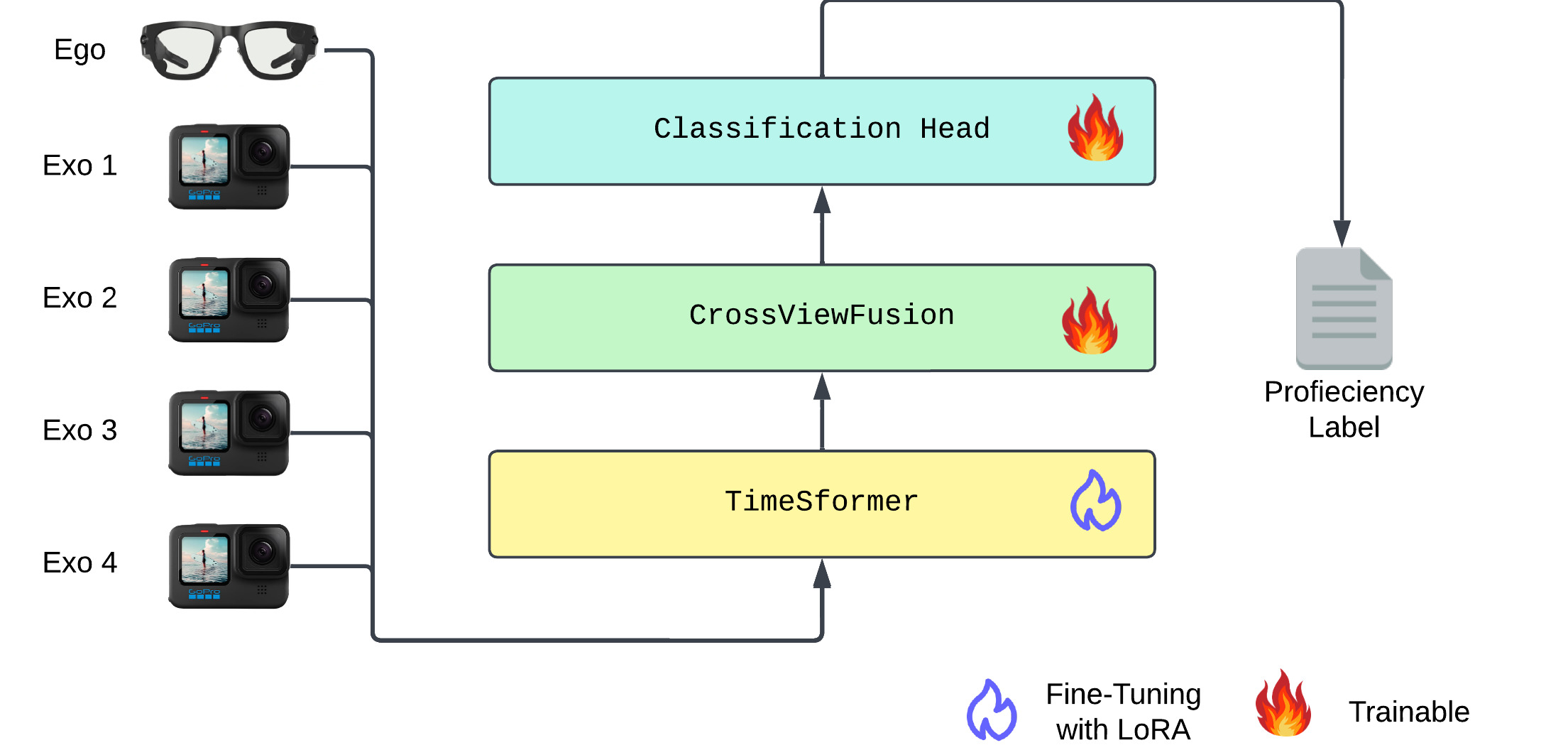}\\[1pt]
    {\scriptsize (a) SkillFormer}
  \end{minipage}\hfill
  \begin{minipage}[c]{0.49\linewidth}
    \centering
    \includegraphics[width=\linewidth]{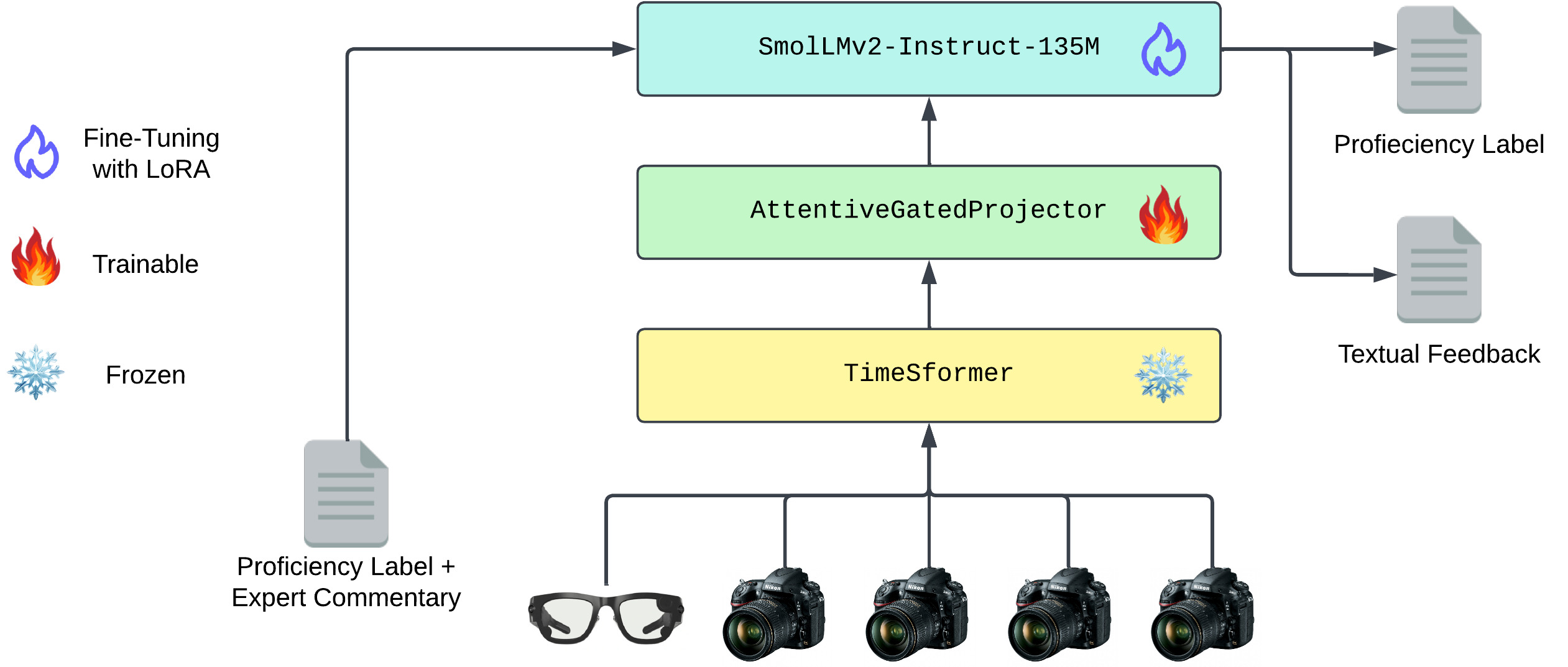}\\[1pt]
    {\scriptsize (b) ProfVLM}
  \end{minipage}
  \caption{End-to-end architectures, both built on a TimeSformer
  backbone. (a) SkillFormer: LoRA-adapted backbone, CrossViewFusion,
  classification head. (b) ProfVLM: frozen backbone, AGP projector
  into a LoRA-adapted SmolLM2-135M producing label and feedback.}
  \label{fig:archs}
\end{figure}

\begin{figure}[t]
  \centering
  \begin{minipage}[c]{0.66\linewidth}
    \centering
    \includegraphics[width=\linewidth]{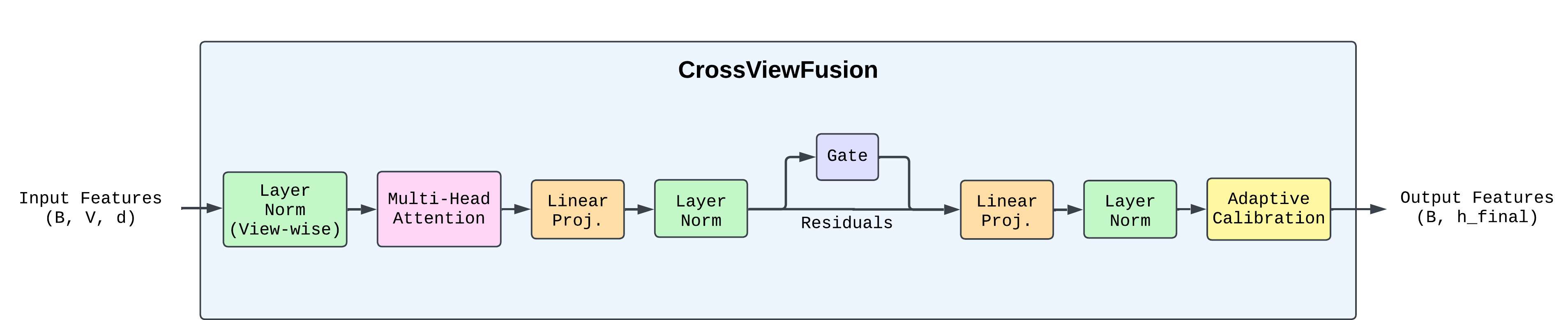}\\[1pt]
    {\scriptsize (a) CrossViewFusion (SkillFormer)}
  \end{minipage}\hfill
  \begin{minipage}[c]{0.30\linewidth}
    \centering
    \includegraphics[width=\linewidth]{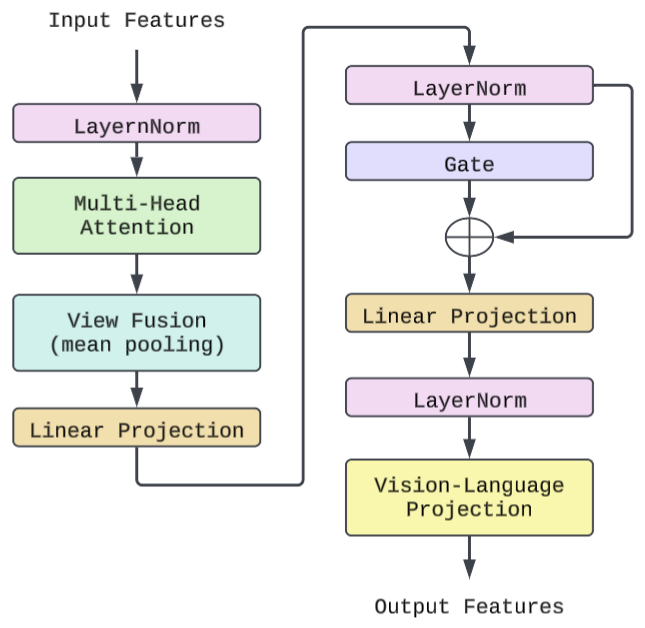}\\[1pt]
    {\scriptsize (b) AGP (ProfVLM)}
  \end{minipage}
  \caption{Multi-view fusion modules. (a) CrossViewFusion: multi-head
  cross-attention, per-view scalar gates, adaptive self-calibration.
  (b) AGP: cross-view attention, mean-pooled fusion, per-token
  sigmoid gate, projection into the language-backbone embedding.}
  \label{fig:projectors}
\end{figure}

SkillFormer~\cite{10.1117/12.3093974} (Fig.~\ref{fig:archs}\,(a))
encodes each of $V$ synchronised views (one egocentric and four
exocentric) with a shared TimeSformer~\cite{timesformer} backbone
pretrained on Kinetics-600 \cite{kinetics}. The backbone is adapted with
LoRA~\cite{lora} on the attention projections, output layers,
temporal-attention components, and feed-forward layers, yielding
$14$--$27$M trainable parameters, depending on rank and scaling configuration. View-specific embeddings are fused by
CrossViewFusion (Fig.~\ref{fig:projectors}\,(a)): view-wise
normalisation and multi-head cross-view attention are followed by
mean aggregation, a feed-forward transformation, an element-wise
learnable gate, and adaptive self-calibration with learnable
feature-wise statistics. The reported configurations use $32$ frames
for Ego, $24$ for Exos, and $16$ for Ego+Exos, with increasing LoRA
rank and fusion capacity as the number of views grows. Trained for 4 epochs, SkillFormer surpasses the Ego-Exo4D multi-view baselines
with $4.5\times$ fewer trainable parameters and a $3.75\times$
shorter training (Tables \ref{tab:overall}, \ref{tab:scenario}).

\subsection{PATS: Proficiency-Aware Temporal Sampling}
\label{sec:pats}
Uniform sampling spreads a fixed frame budget across the whole clip,
providing broad coverage but low local temporal density. This can miss
the evolution of \emph{fundamental movements} through which
proficiency is expressed, such as a shot, a climbing move, or a
musical phrase. PATS~\cite{pats} addresses this by concentrating
frames within short, continuous action segments while still sampling
multiple parts of the video. Given $N_{\text{target}}$ frames, it
selects $N_s$ continuous temporal segments of duration $d_s$,
distributes the frame budget across them, and samples densely within
each segment. Segment starts are spread over the video to retain
coverage, while segment duration is shortened when needed to avoid
overlap.

PATS is architecture-agnostic: it replaces SkillFormer's sampler
without changing the model or training setup. This improves all
Ego-Exo4D view configurations, reaching $47.3\%$ for Ego, $46.6\%$
for Exos, and $48.0\%$ for Ego+Exos (Table~\ref{tab:overall}). The
largest gains occur in domains where skill depends on temporally
coherent movement patterns, such as bouldering, music, and basketball
(Table~\ref{tab:scenario}).

\subsection{ProfVLM: From Classification to Generative Feedback}
\label{sec:profvlm}
ProfVLM~\cite{BIANCHI2026104749} (Fig.~\ref{fig:archs}\,(b)) is the
first vision--language model for multi-view proficiency estimation
that predicts skill entirely through \emph{conditional language
generation}, without a dedicated classification head. A single
autoregressive output contains both the proficiency level and
natural-language feedback. A frozen TimeSformer~\cite{timesformer},
pretrained on Kinetics-600~\cite{kinetics}, encodes 8-frame clips from
each view. The AttentiveGatedProjector (AGP,
Fig.~\ref{fig:projectors}\,(b)) normalises view-specific features,
fuses them with multi-head cross-view attention and mean pooling, and
aligns the fused representation with the language-model embedding
space through feed-forward refinement, element-wise gating,
projection, and learned normalisation. The resulting embeddings are
inserted as special video tokens into
SmolLM2-135M-Instruct~\cite{smollm2}, which is LoRA-adapted for
generation. Trained on Ego-Exo4D videos and expert commentaries with a
causal language-modelling objective, ProfVLM generates outputs of the
form ``\texttt{Proficiency Level: <label>; Proficiency Commentary:
<feedback>}'', from which the label is parsed. With only $5.3$M
trainable parameters, 8 input frames, and 6 training epochs,
ProfVLM reaches $48.2\%$ top-1 accuracy on Ego$+$Exos, surpassing
SkillFormer while using about $5\times$ fewer trainable parameters that SkillFormer 
and $20\times$ fewer than TimeSformer baselines (Tables \ref{tab:overall}, \ref{tab:scenario}).

\begin{table}[t]
  \caption{Top-1 accuracy (\%) on Ego-Exo4D. The rightmost column
  distinguishes discriminative classifiers, which predict a label through
  a classification head, from generative models, which produce the label
  as text. \textbf{Bold}: best; \underline{underline}: second-best.}
  \label{tab:overall}
  \centering\footnotesize
  \setlength{\tabcolsep}{4pt}
  \renewcommand{\arraystretch}{1.0}
  \begin{tabular}{lccccccc}
    \toprule
    Method & Ego & Exos & Ego$+$Exos & Params & Frames & Epochs & Paradigm \\
    \midrule
    Random
      & 24.9 & 24.9 & 24.9 & -- & -- & -- 
      & \multirow{2}{*}{\scriptsize --} \\
    Majority
      & 31.1 & 31.1 & 31.1 & -- & -- & --
      & \\
    \midrule
    Ego-Exo4D Baselines (TimeSformer)~\cite{egoexo4d}
      & 46.8 & 40.6 & 40.8 & 121M & 16 & 15
      & \multirow{4}{*}{\scriptsize\begin{tabular}{@{}l@{}}Discriminative\\(Classification)\end{tabular}} \\
    EgoPulseFormer~\cite{egoppg}
      & 45.3 & 35.9 & 42.4 & 121M & 16 & 15
      & \\
    SkillFormer~\cite{10.1117/12.3093974}
      & 45.9 & \underline{46.3} & 47.5 & 27M & 16--32 & 4
      & \\
    SkillFormer$+$PATS~\cite{pats}
      & \underline{47.3} & 46.6 & \underline{48.0} & 27M & 24--32 & 4
      & \\
    \midrule
    ProfVLM (AGP)~\cite{BIANCHI2026104749}
      & 44.2 & 45.1 & \textbf{48.2} & 5.3M & 8 & 6
      & {\scriptsize Generative} \\
    \bottomrule
  \end{tabular}
\end{table}

\begin{table}[t]
  \caption{Per-scenario top-1 accuracy (\%) by view configuration. \textbf{Bold}: best per scenario;
  \underline{underline}: second-best.}
  \label{tab:scenario}
  \centering\footnotesize
  \setlength{\tabcolsep}{4pt}
  \renewcommand{\arraystretch}{1.0}
  \begin{tabular}{llccccccc}
    \toprule
    Method & View & Basket. & Cook. & Dance & Music & Bould. & Soccer & Overall \\
    \midrule
    \multirow{3}{*}{EgoExo4D Baseline \cite{egoexo4d}}
      & Ego        & 51.4 & 45.0 & \underline{55.7} & 46.2 & 25.3 & 56.3 & 46.8 \\
      & Exos       & 52.3 & 35.0 & 42.7 & 69.2 & 17.3 & \underline{75.0} & 40.6 \\
      & Ego$+$Exos & 55.2 & 35.0 & 42.7 & 56.4 & 17.3 & \underline{75.0} & 40.8 \\
    \midrule
    \multirow{3}{*}{SkillFormer \cite{10.1117/12.3093974}}
      & Ego        & 69.0 & 31.6 & 20.5 & \underline{72.4} & 30.8 & 70.8 & 45.9 \\
      & Exos       & 70.8 & 47.4 & 15.4 & 69.0 & 33.5 & 66.7 & 46.3 \\
      & Ego$+$Exos & \underline{77.9} & \textbf{60.5} & 13.7 & 68.1 & 31.9 & 66.7 & 47.5 \\
    \midrule
    \multirow{3}{*}{SkillFormer$+$PATS \cite{pats}}
      & Ego        & 64.6 & 39.5 & 22.2 & \textbf{74.1} & \textbf{42.3} & 66.7 & 47.3 \\
      & Exos       & 72.6 & \textbf{60.5} & 20.5 & 69.8 & 36.8 & 66.7 & 46.6 \\
      & Ego$+$Exos & \textbf{78.8} & 50.1 & 26.5 & 69.0 & 36.3 & 66.7 & \underline{48.0} \\
    \midrule
    \multirow{3}{*}{ProfVLM \cite{BIANCHI2026104749}}
      & Ego        & 36.0 & 31.0 & 51.4 & 72.1 & 37.5 & 57.3 & 44.2 \\
      & Exos       & 33.0 & \underline{56.0} & 53.9 & 61.5 & 37.5 & \textbf{76.0} & 45.1 \\
      & Ego$+$Exos & 41.0 & 51.0 & \textbf{60.4} & 56.3 & \underline{38.7} & 69.8 & \textbf{48.2} \\
    \bottomrule
  \end{tabular}
\end{table}

\begin{table}[t]
  \caption{Quality of the natural-language feedback produced by
  ProfVLM~\cite{BIANCHI2026104749} (AGP variant) across view
  configurations. No prior work in proficiency estimation reports a
  comparable evaluation.}
  \label{tab:nlg}
  \centering\footnotesize
  \setlength{\tabcolsep}{6pt}
  \renewcommand{\arraystretch}{1.02}
  \begin{tabular}{lccc}
    \toprule
    View & BERTScore (F1) & METEOR & ROUGE-L \\
    \midrule
    Ego        & 85.41 & 18.06 & 14.47 \\
    Exos       & 85.51 & 17.33 & 15.67 \\
    Ego$+$Exos & \textbf{85.53} & \textbf{18.23} & \textbf{15.65} \\
    \bottomrule
  \end{tabular}
\end{table}

\section{Discussion}
\label{sec:disc}

The results in Tables~\ref{tab:overall}--\ref{tab:nlg} point to four
main design lessons: selective view fusion, temporal sampling,
generative output, and domain-aware adaptation.

\paragraph{View selection and fusion.}
Adding views is not sufficient by itself. In the Ego-Exo4D baselines,
the best TimeSformer Ego result is $46.8\%$, while Ego$+$Exos drops
to $40.8\%$, indicating that unstructured fusion can dilute useful
cues. The per-scenario results confirm that the best viewpoint is
domain-dependent (Table~\ref{tab:scenario}). SkillFormer addresses
this with CrossViewFusion, reaching $47.5\%$ on Ego$+$Exos with
$4.5\times$ fewer trainable parameters than the TimeSformer baselines;
ProfVLM's AGP raises the combined setting to $48.2\%$. Thus,
the key issue is not view availability, but view alignment and fusion.

\paragraph{Frames and temporal sampling.}
More frames do not automatically improve proficiency estimation. The
models that use fewer frames can match or surpass heavier baselines
when temporal information is sampled and fused more effectively:
ProfVLM reaches the best Ego$+$Exos result with only 8 frames, while
SkillFormer and PATS use 16--32 frames (Table~\ref{tab:overall}).
Multi-view input can also compensate for shorter clips, provided
that the views are aligned and selectively fused. PATS shows
that the temporal sampling pattern matters: by increasing local
sampling density within continuous segments, it improves SkillFormer
in all view configurations and yields the largest gains in domains
with structured fundamental movements, such as bouldering, music, and
basketball (Table~\ref{tab:scenario}).

\paragraph{From classification to generation.}
ProfVLM replaces the classification head with a language model that
produces a structured \texttt{Level+Feedback} response, from which the
label is parsed deterministically. This slightly surpasses
SkillFormer$+$PATS on Ego$+$Exos ($48.2\%$ vs. $48.0\%$;
Table~\ref{tab:overall}) while using roughly one fifth of the
trainable parameters. It also generates expert-style feedback
(Table~\ref{tab:nlg}), adding interpretability without an accuracy
penalty.

\paragraph{Domain-aware adaptation.}
Per-domain results remain heterogeneous (Table~\ref{tab:scenario}).
PATS shows that there is no single temporal configuration that is
optimal for all activities: domains differ in the useful view, the
preferred sampling density, and the amount of temporal continuity they
require. This suggests shared visual encoders with lightweight
domain-specific adapters or sampling policies, rather than a single
monolithic model for all skills.

\section{Conclusions and Outlook}
\label{sec:conclusions}
SkillFormer, PATS, and ProfVLM jointly advance multi-view proficiency
estimation on Ego-Exo4D with substantially reduced trainable-parameter
budgets. Together, they shift the design space from closed-set
classification toward systems that combine selective view fusion,
smart temporal sampling, and generative expert-style
feedback. The frozen-backbone, AGP, and compact-LM stack used by
ProfVLM is compatible with video-LLM agent orchestration~\cite{videoagent},
opening the way to interactive systems that observe an athlete across
sessions and adapt their feedback over time. Another natural direction
is to add structured motion cues: Gate-Shift-Pose~\cite{gsp} suggests
that explicit pose information can help when motion quality is
discriminative. Beyond reducing trainable parameters with LoRA and lightweight
projectors, KD-AHOSVD~\cite{kdohsvd} and related plug-and-play KD
modules~\cite{kdohsvd-paper} could further compress the overall models
and support on-device deployment. Evaluation remains equally important:
future benchmarks should combine multi-view recordings, expert
critiques, and human ratings of feedback actionability, while
accounting for long-term adaptation, personalisation, and privacy.

\section*{Declaration on Generative AI}
  The author(s) have not employed any Generative AI tools.


\bibliography{sample-ceur}

@String{Computer = "{IEEE} Computer" }

@String{Springer = "Springer-Verlag" }

@article{BIANCHI2026104749,
  author  = {Edoardo Bianchi and Jacopo Staiano and Antonio Liotta},
  title   = {{ProfVLM}: A lightweight video-language model for multi-view proficiency estimation},
  journal = {Computer Vision and Image Understanding},
  volume  = {268},
  pages   = {104749},
  year    = {2026},
  doi     = {10.1016/j.cviu.2026.104749},
}

@inproceedings{10.1117/12.3093974,
  author    = {Edoardo Bianchi and Antonio Liotta},
  title     = {{SkillFormer}: unified multiview video understanding for proficiency estimation},
  booktitle = {Eighteenth International Conference on Machine Vision (ICMV 2025)},
  volume    = {14114},
  pages     = {141142G},
  year      = {2026},
  publisher = {SPIE},
  doi       = {10.1117/12.3093974},
}

@INPROCEEDINGS{pats,
  author    = {Bianchi, Edoardo and Liotta, Antonio},
  booktitle = {2025 IEEE International Workshop on Sport, Technology and Research (STAR)},
  title     = {{PATS}: Proficiency-Aware Temporal Sampling for Multi-View Sports Skill Assessment},
  year      = {2025},
  pages     = {1--6},
  doi       = {10.1109/STAR66750.2025.11264769},
}

@InProceedings{gsp,
  author    = {Bianchi, Edoardo and Lanz, Oswald},
  title     = {Gate-Shift-Pose: Enhancing Action Recognition in Sports with Skeleton Information},
  booktitle = {Proceedings of the Winter Conference on Applications of Computer Vision (WACV) Workshops},
  year      = {2025},
  pages     = {1257--1264},
}

@inproceedings{gsfmeccano,
  author    = {Bianchi, Edoardo and Lanz, Oswald},
  title     = {Egocentric Video-Based Human Action Recognition in Industrial Environments},
  booktitle = {Latest Advancements in Mechanical Engineering},
  publisher = {Springer Nature Switzerland},
  year      = {2024},
  pages     = {257--267},
}

@inproceedings{kdohsvd,
  author    = {Meneghetti, Laura and Bianchi, Edoardo and Demo, Nicola and Rozza, Gianluigi},
  title     = {{KD-AHOSVD}: Neural Network Compression via Knowledge Distillation and Tensor Decomposition},
  booktitle = {Design and Architecture for Signal and Image Processing},
  publisher = {Springer Nature Switzerland},
  year      = {2025},
  pages     = {81--92},
}

@article{kdohsvd-paper,
	author = {Laura Meneghetti and Edoardo Bianchi and Nicola Demo and Gianluigi Rozza},
	doi = {10.1016/j.sysarc.2026.103778},
	journal = {Journal of Systems Architecture},
	pages = {103778},
	title = {Plug-and-play neural compression: A knowledge distillation framework with flexible dimensionality reduction},
	volume = {175},
	year = {2026}}

@InProceedings{egoexo4d,
  author    = {Grauman, Kristen and Westbury, Andrew and Torresani, Lorenzo and Kitani, Kris and Malik, Jitendra and others},
  title     = {Ego-Exo4D: Understanding Skilled Human Activity from First- and Third-Person Perspectives},
  booktitle = {Proceedings of the IEEE/CVF Conference on Computer Vision and Pattern Recognition (CVPR)},
  year      = {2024},
  pages     = {19383--19400},
}

@misc{AQA_survey,
  author        = {Kanglei Zhou and Ruizhi Cai and Liyuan Wang and Hubert P. H. Shum and Xiaohui Liang},
  title         = {A Comprehensive Survey of Action Quality Assessment: Method and Benchmark},
  year          = {2024},
  eprint        = {2412.11149},
  archivePrefix = {arXiv},
}

@INPROCEEDINGS{parmar2019mtl,
  author    = {Parmar, Paritosh and Morris, Brendan Tran},
  booktitle = {IEEE/CVF Conference on Computer Vision and Pattern Recognition (CVPR)},
  title     = {What and How Well You Performed? A Multitask Learning Approach to Action Quality Assessment},
  year      = {2019},
  pages     = {304--313},
  doi       = {10.1109/CVPR.2019.00039},
}

@INPROCEEDINGS{zhang2024nae,
  author    = {Zhang, Shiyi and Bai, Sule and Chen, Guangyi and Chen, Lei and Lu, Jiwen and Wang, Junle and Tang, Yansong},
  booktitle = {IEEE/CVF Conference on Computer Vision and Pattern Recognition (CVPR)},
  title     = {Narrative Action Evaluation with Prompt-Guided Multimodal Interaction},
  year      = {2024},
  pages     = {18430--18439},
  doi       = {10.1109/CVPR52733.2024.01744},
}

@misc{egoppg,
  author        = {Björn Braun and Rayan Armani and Manuel Meier and Max Moebus and Christian Holz},
  title         = {egoPPG: Heart Rate Estimation from Eye-Tracking Cameras in Egocentric Systems to Benefit Downstream Vision Tasks},
  year          = {2025},
  eprint        = {2502.20879},
  archivePrefix = {arXiv},
}

@InProceedings{pan2025basket,
  author    = {Pan, Yulu and Zhang, Ce and Bertasius, Gedas},
  title     = {{BASKET}: A Large-Scale Video Dataset for Fine-Grained Skill Estimation},
  booktitle = {Proceedings of the IEEE/CVF Conference on Computer Vision and Pattern Recognition (CVPR)},
  year      = {2025},
  pages     = {28952--28962},
}

@misc{timesformer,
  author        = {Gedas Bertasius and Heng Wang and Lorenzo Torresani},
  title         = {Is Space-Time Attention All You Need for Video Understanding?},
  year          = {2021},
  eprint        = {2102.05095},
  archivePrefix = {arXiv},
}

@inproceedings{lora,
  author    = {Edward J. Hu and Yelong Shen and Phillip Wallis and Zeyuan Allen-Zhu and Yuanzhi Li and Shean Wang and Lu Wang and Weizhu Chen},
  title     = {{LoRA}: Low-Rank Adaptation of Large Language Models},
  booktitle = {International Conference on Learning Representations (ICLR)},
  year      = {2022},
}

@misc{smollm2,
  author        = {Loubna Ben Allal and Anton Lozhkov and Elie Bakouch and others},
  title         = {{SmolLM2}: When Smol Goes Big -- Data-Centric Training of a Small Language Model},
  year          = {2025},
  eprint        = {2502.02737},
  archivePrefix = {arXiv},
}

@inproceedings{llava,
  author    = {Liu, Haotian and Li, Chunyuan and Wu, Qingyang and Lee, Yong Jae},
  title     = {Visual Instruction Tuning},
  booktitle = {NeurIPS},
  year      = {2023},
}

@misc{videoagent,
  author        = {Xiaohan Wang and Yuhui Zhang and Orr Zohar and Serena Yeung-Levy},
  title         = {{VideoAgent}: Long-form Video Understanding with Large Language Model as Agent},
  year          = {2024},
  eprint        = {2403.10517},
  archivePrefix = {arXiv},
}

@misc{tacticexpert,
  author        = {Xu Lingrui and Liu Mandi and Zhang Lei},
  title         = {{TacticExpert}: Spatial-Temporal Graph Language Model for Basketball Tactics},
  year          = {2025},
  eprint        = {2503.10722},
  archivePrefix = {arXiv},
}

@misc{kinetics,
      title={The Kinetics Human Action Video Dataset}, 
      author={Will Kay and Joao Carreira and Karen Simonyan and Brian Zhang and Chloe Hillier and Sudheendra Vijayanarasimhan and Fabio Viola and Tim Green and Trevor Back and Paul Natsev and Mustafa Suleyman and Andrew Zisserman},
      year={2017},
      eprint={1705.06950},
      archivePrefix={arXiv},
      primaryClass={cs.CV},
      url={https://arxiv.org/abs/1705.06950}, 
}

\appendix

\end{document}